\documentclass[runningheads]{llncs}
\usepackage[T1]{fontenc}
\usepackage{amsfonts,amssymb}
\usepackage{multirow}
\usepackage{amsmath}
\usepackage{graphicx}
\usepackage{subcaption}
\usepackage{mathrsfs}
\usepackage{caption}
\usepackage{amssymb} 
\usepackage{booktabs}
\usepackage{hyperref}
\usepackage{colortbl}
\usepackage{makeidx}
\usepackage{tabularx}
\makeindex
\DeclareCaptionLabelSeparator{dot}{. }
\usepackage{siunitx} 
\usepackage[table]{xcolor} 
\usepackage{subcaption}
\usepackage{xcolor}
\usepackage{algorithm}
\usepackage{algpseudocode}
\usepackage[misc]{ifsym}
\usepackage{booktabs}
\hypersetup{
  colorlinks=true,
  linkcolor=blue,
  citecolor=blue, 
  urlcolor=blue, 
  linkbordercolor=white, 
  citebordercolor=white
}
\definecolor{lightyellow}{rgb}{1.0, 1.0, 0.6}
\begin{document}

\title{Hierarchical Semantic Learning for Multi-Class Aorta Segmentation}

\author{Pengcheng Shi\inst{1,2}\orcidID{0000-0003-2180-2312}}


\authorrunning{P. Shi et al.}

\institute{Medical Image Insights, Foundation Models Division, Shanghai, China\\ \email{shipc1220@gmail.com} \and Harbin Institute of Technology (Shenzhen), Shenzhen, China}

\maketitle 

\begin{abstract}
The aorta, the body's largest artery, is prone to pathologies such as dissection, aneurysm, and atherosclerosis, which often require timely intervention. Minimally invasive repairs involving branch vessels necessitate detailed 3D anatomical analysis. Existing methods often overlook hierarchical anatomical relationships while struggling with severe class imbalance inherent in vascular structures. We address these challenges with a curriculum learning strategy that leverages a novel \textbf{fractal softmax} for hierarchical semantic learning. Inspired by human cognition, our approach progressively learns anatomical constraints by decomposing complex structures from simple to complex components. The curriculum learning framework naturally addresses class imbalance by first establishing robust feature representations for dominant classes before tackling rare but anatomically critical structures, significantly accelerating model convergence in multi-class scenarios. Our two-stage inference strategy achieves up to fivefold acceleration, enhancing clinical practicality. On the validation set at epoch 50, our hierarchical semantic loss improves the Dice score of nnU-Net ResEnc M by 11.65\%. The proposed model demonstrates a 5.6\% higher Dice score than baselines on the test set. Experimental results show significant improvements in segmentation accuracy and efficiency, making the framework suitable for real-time clinical applications. The implementation code for this challenge entry is publicly available at: \url{https://github.com/PengchengShi1220/AortaSeg24}. The code for fractal softmax will be available at \url{https://github.com/PengchengShi1220/fractal-softmax}.

\end{abstract}

\section{Introduction}
Minimally invasive aortic repairs require precise 3D anatomical analysis, which can be achieved through multi-class segmentation of the aorta in CTA to facilitate appropriate device selection. However, current segmentation methods face two fundamental challenges: (1) they often ignore the hierarchical anatomical information of vascular structures, which is essential for preserving anatomical consistency, and (2) they struggle with severe class imbalance inherent in vascular networks, where major vessels dominate while critical branch structures appear only sparsely.

The challenge of class imbalance in tubular structure segmentation is particularly pronounced, where the volume ratio between dominant classes and fine peripheral vessels can be extremely large. Conventional approaches like class reweighting or focal loss~\cite{lin2017focal} often fail to fully utilize the natural semantic hierarchy. Instead, our reference to the hierarchical semantic segmentation network (HSSN)~\cite{li2023semantic} addresses this imbalance by incorporating structural curriculum learning, where the model first learns to distinguish merged large-volume classes before progressing to finer branch-level distinctions.

We introduce a novel fractal softmax module that leverages curriculum learning to drive a hierarchical framework for multi-class aorta segmentation. Curriculum learning, where examples are introduced progressively from simple to complex \cite{bengio2009curriculum}, provides a natural solution to class imbalance by first establishing robust feature representations for dominant structures before tackling rare but anatomically critical branches. This approach mirrors the hierarchical learning process observed in human cognition \cite{wang2021survey}, allowing the model to find a more optimal parameter space while simultaneously addressing the class imbalance problem. The combination of hierarchical semantics with curriculum learning proves particularly effective for vascular segmentation, where the progressive learning strategy significantly accelerates convergence compared to flat learning paradigms, especially when dealing with numerous anatomical classes. Our model decomposes anatomical semantics from simpler to more complex structures, utilizing a hierarchical semantic tree to ensure consistent and anatomically sound segmentation while explicitly modeling the relative prevalence of different vascular classes.

The integration of the cbDice loss function further ensures uniform segmentation across classes while maintaining the vascular network's topological integrity, with the boundary-aware formulation naturally compensating for class imbalance through geometric constraints rather than simple frequency-based reweighting. The two-stage inference method first segments all foreground vessels using a low-resolution 3D model, followed by multi-class segmentation in a refined region of interest (ROI). This hierarchical processing not only reduces inference time by up to 5x but also helps mitigate class imbalance by focusing computational resources on regions containing minority classes.

We extend these concepts by incorporating a hierarchical semantic tree that explicitly models prevalence relationships between classes, allowing the network to leverage the natural anatomical hierarchy when learning rare classes. The hierarchical constraints also serve as a form of data augmentation for minority classes, as correctly predicting their parent classes in the hierarchy provides strong spatial priors for their location.

\subsection{Hierarchical Semantic Learning}
Hierarchical semantic learning (HSL) addresses the inherent limitations of conventional pixel-wise segmentation methods by incorporating structured class relationships into the learning process. Unlike flat segmentation approaches that enforce only fine-grained semantic constraints \cite{isensee2021nnu}, HSL leverages multi-level semantic hierarchies to improve feature discrimination, particularly in homogeneous multi-class tubular structures. This paradigm aligns with human cognition, where understanding progresses from coarse classes to fine-grained classes.

\subsection{Curriculum Learning}
Curriculum learning (CL) progressively introduce samples of increasing complexity. Inspired by human educational strategies \cite{bengio2009curriculum}, CL trains model on easier subsets (e.g., larger organs or isolated structures) before advancing to finer, harder-to-segment classes (e.g., small vessels or overlapping airways). This staged approach is particularly effective for multi-class datasets \cite{isensee2021nnu,wasserthal2023totalsegmentator}, where early exposure to dominant classes can stabilize feature learning before addressing rare or morphologically complex targets. However, current CL implementations often decouple semantic hierarchy from the curriculum, limiting their ability to exploit structural relationships. Integrating HSL with CL—for instance, by sequencing tasks from coarse-to-fine hierarchies—could further enhance model robustness in heterogeneous anatomical segmentation.

\begin{figure}[t]
\centering
\includegraphics[width=\textwidth]{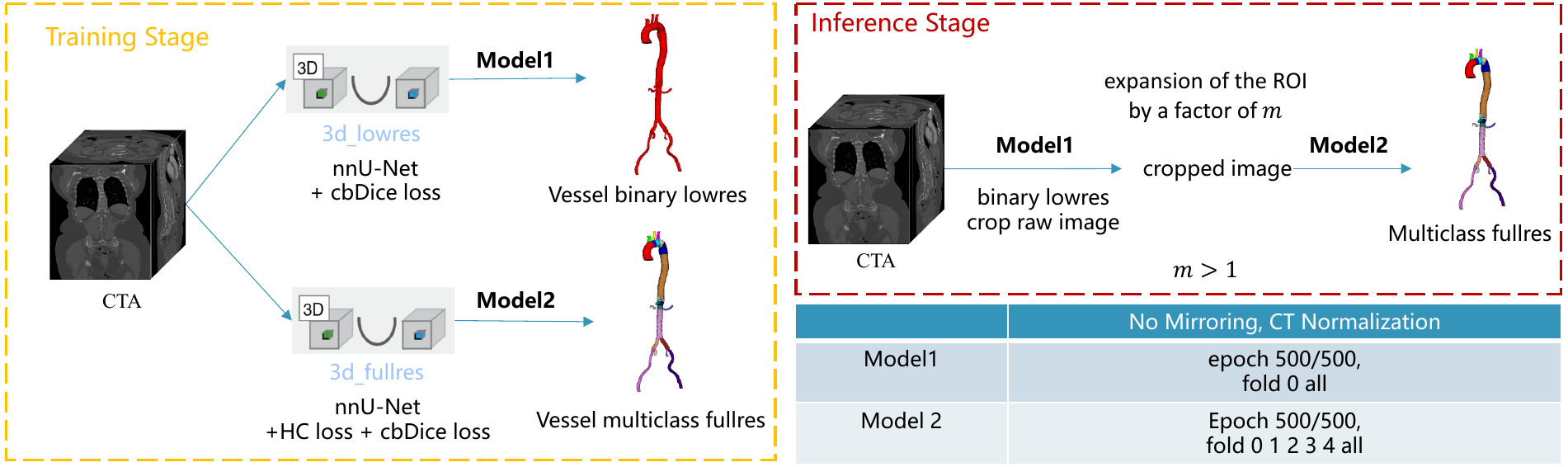}
\caption{The two-stage inference approach improves speed by up to 5 times compared to using only Model 2, depending on the expansion factor $m$ of the ROI in the first stage.}
\label{fig1}
\end{figure}

\section{Methodology}
\subsection{Fractal Softmax}
Drawing inspiration from the HSSN \cite{li2023semantic}, we focus on the hierarchical interrelationships among classes using a tree structure \( \mathcal{T} = (\mathcal{V}, \mathcal{E}) \). Each node \( v \in \mathcal{V} \) symbolizes a semantic (or anatomical) category, with edges \( (u, v) \in \mathcal{E} \) denoting hierarchical relationships, where node \( v \) is the superclass of \( u \). Layers of tree \( \mathcal{T} \) are denoted by \( l \) (where \( l = 1, 2, \ldots, L \)), with \( \mathcal{V}_l \) representing the set of all foreground categories at level \( l \).

For volumetric segmentation, assuming that the 3D input (volume) $\textbf{x}\in\mathbb{R}^{C\times H\times W\times D}$ with resolution $(H,W,D)$, $C$ base channels. A segmentation network \( f_{\text{NET}} \) is devised to extract the finest-level leaf predictions \( \textbf{y}_{L} = f_{\text{NET}}(\textbf{x}) \), where \( \textbf{y}_L \in \mathbb{R}^{(|\mathcal{V}_{L}|+1) \times H \times W \times D} \).

In addition to the finest level of leaf predictions, for levels \( l \in \{L-1, L-2, \ldots, 1\} \), our model predicts semantic categories \( v_l \) for each pixel \( y \) by recursively computing \( y_{v_l} = \max_{u \in \mathcal{C}_{v_l}} y_u \). The predictions at each hierarchical level \( l \) are represented as \( \textbf{y}_l \), where each class \( v_l \) at level \( l \) is expressed by \( \textbf{y}_l[v_l] \in \mathbb{R}^{H \times W \times D} \). The set \( \mathcal{C}_{v_l} \) encompasses subclasses of node \( v_l \), and \( \mathcal{A}_{v_l} \) its superclasses. We apply softmax to \( \textbf{y}_l \) to obtain probabilistic predictions \( \textbf{p}_l \), The process is shown in Algorithm~\ref{alg:hierarchical_softmax}.

\begin{algorithm}
\caption{Fractal softmax prediction}
\label{alg:hierarchical_softmax}
\begin{algorithmic}[1]
\Require Image \( \textbf{x} \), Tree structure \( \mathcal{T} = (\mathcal{V}, \mathcal{E}) \)
\Ensure Hierarchical semantic predictions from the finest (leaf) to the coarsest (root) level
\State Compute initial leaf predictions \( \textbf{y}_{L} = f_{\text{NET}}(\textbf{x}) \)  
\For{\( l = L-1 \) to \( 1 \)} 
    \State Initialize \( \textbf{y}_l \) as an empty tensor of appropriate dimensions
    \For{each node \( v_l \in \mathcal{V}_l \)}  
        \State Compute subclass set \( \mathcal{C}_{v_l} \)  
        \State \( \textbf{y}_l[v_l] \gets \max_{u \in \mathcal{C}_{v_l}} \textbf{y}_{l+1}[u] \)  
    \EndFor
    \State \( \textbf{p}_l \gets \text{softmax}(\textbf{y}_l) \)  
\EndFor
\State \Return \( \{\textbf{p}_l\}_{l=1}^{L} \)  
\end{algorithmic}
\end{algorithm}

In the multi-class scenario, the hierarchical tree \(\mathcal{T}\) involves various subclasses and superclasses at different levels, as well as sibling categories at the same level, all of which need to fulfill specific properties and constraints. Some properties have been explored in various hierarchical classification works \cite{bi2011multi,wehrmann2018hierarchical,giunchiglia2020coherent}, though some aspects have received less attention. In the context of the HSSN, the hierarchical tree \(\mathcal{T}\) is governed by six essential properties or constraints that influence the behavior of the segmentation model (Definitions 1 to 6).

\textbf{Definition 1} (Positive \(\mathcal{T}\)-Property). \textit{If a category is marked positive for a pixel, all its ancestor nodes in \( \mathcal{T} \) (i.e., superclasses) are also positive.}

\textbf{Definition 2} (Negative \(\mathcal{T}\)-Property). \textit{If a category is marked negative for a pixel, all its descendant nodes in \( \mathcal{T} \) (i.e., subclasses) are also negative.}

\textbf{Definition 3} (Positive \(\mathcal{T}\)-Constraint). \textit{For a positive marked category \( v \) and its ancestor \( u \), it must hold that \( p_v \leq p_u \).}

\textbf{Definition 4} (Negative \(\mathcal{T}\)-Constraint). \textit{For a negative marked category \( v \) and its descendant \( u \), it must hold that \( 1 - p_v \leq 1 - p_u \).}

\textbf{Definition 5} (Exclusivity \( \mathcal{T} \)-Property). \textit{If a category at a certain hierarchical level is marked positive, all its sibling nodes (categories at the same level) in \( \mathcal{T} \) should be marked negative.}

\textbf{Definition 6} (Exclusivity \( \mathcal{T} \)-Constraint). \textit{For a positive marked category \( v \) and its sibling \( u \), it should hold that \( p_v \leq 1 - p_u \).}

\begin{figure}[t]
    \centering
    \includegraphics[width=\textwidth]{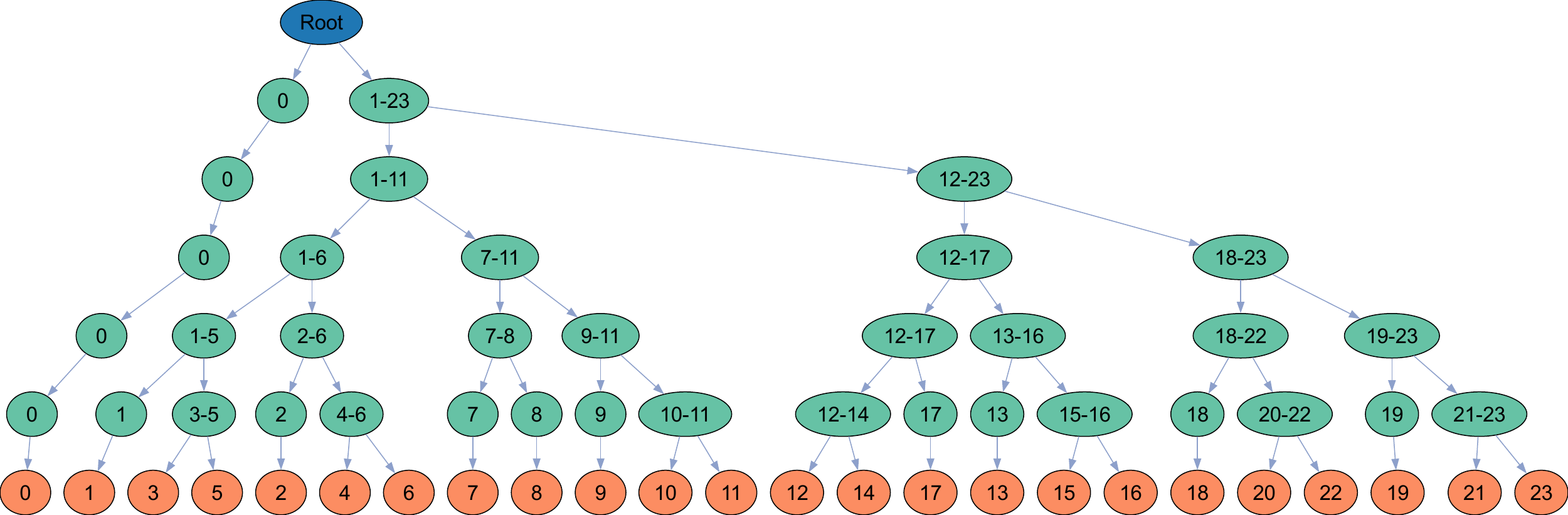}
    \caption{Hierarchical anatomical tree for AortaSeg24, illustrating superclass-subclass relationships from simple to complex structures.}
    \label{fig:aorta_tree}
\end{figure}

Unlike the HSSN, which employs sigmoid functions for the hierarchical classification of each category \( v \in \mathcal{V} \), our method adopts the softmax function to normalize prediction results \(\textbf{y}_l\) at each hierarchical level \(l\). We introduce this as the "fractal softmax" in this paper. It shares certain similarities with the hierarchical recursive generation process of fractal models \cite{li2025fractal}. This approach is designed to satisfy both the positive and negative \(\mathcal{T}\)-properties and constraints, as well as the exclusivity properties and constraints, as outlined in Definitions 1 through 6. Figure~\ref{fig:aorta_tree} illustrates the hierarchical semantic tree for the AortaSeg24 dataset, which includes 23 classes. The tree structure captures the relationships between superclasses and subclasses, facilitating the hierarchical learning process.

\subsection{Two-Stage Inference}
As shown in Fig.~\ref{fig1}, the training framework is based on nnU-Net V2 \cite{isensee2021nnu}. To enhance efficiency, we adopt a two-stage approach: first, a low-resolution 3D model segments all foreground vessels, followed by multi-class segmentation within an ROI extracted from the original image. This significantly improves inference speed, making the model more suitable for clinical applications.

\subsection{Training Details}
The cl-X-Dice \cite{shi2024centerline} addresses segmentation challenges for vessels of varying diameters:
\begin{equation}
\label{eq:Tprec_clXDice}
\mathrm{Tprec}(S_{\mathrm{P}}, S_{\mathrm{L}}, V_{\mathrm{L}}) = \frac{|Q_{\mathrm{sp}} \cap Q_{\mathrm{vl}}|}{|Q_{\mathrm{sp}} \cap Q_{\mathrm{spvp}} \cap (U - S_{\mathrm{L}})| + |Q_{\mathrm{sp}} \cap Q_{\mathrm{slvl}}|}
\end{equation}
\begin{equation}
\label{eq:Tsens_clXDice}
\mathrm{Tsens}(S_{\mathrm{L}}, S_{\mathrm{P}}, V_{\mathrm{P}}) = \frac{|Q_{\mathrm{sl}} \cap Q_{\mathrm{vp}}|}{|Q_{\mathrm{sl}} \cap Q_{\mathrm{slvl}} \cap (U - S_{\mathrm{P}})| + |Q_{\mathrm{sl}} \cap Q_{\mathrm{spvp}}|}
\end{equation}
\begin{equation}
\label{eq:cl-X-Dice}
\text{cl-X-Dice}(V_{\mathrm{P}}, V_{\mathrm{L}}) = \frac{2 \times \mathrm{Tprec}(S_{\mathrm{P}}, S_{\mathrm{L}}, V_{\mathrm{L}}) \times \mathrm{Tsens}(S_{\mathrm{L}}, S_{\mathrm{P}}, V_{\mathrm{P}})}{\mathrm{Tprec}(S_{\mathrm{P}}, S_{\mathrm{L}}, V_{\mathrm{L}}) + \mathrm{Tsens}(S_{\mathrm{L}}, S_{\mathrm{P}}, V_{\mathrm{P}})} 
\end{equation}

Model training details:
- \textbf{Model 1:} 500 epochs using low-resolution 3D binary segmentation.
- \textbf{Model 2:} 1000 epochs for full-resolution multi-class segmentation.

The total loss function combines cross-entropy, Dice, and cBDice. The cross-entropy and Dice components are subjected to $L$-level constraints:
\begin{equation}
\label{eq:Loss}
\mathcal{L}(\textbf{y}, \textbf{g}) = \frac{1}{\lambda_{\text{sum}}} \sum_{l=1}^{L_s} \lambda_{l}\cdot\left( \lambda_{\text{ce}}\cdot\mathcal{L}_{\text{ce}}(\textbf{y}_l, \textbf{g}_l) + \lambda_{\text{dice}}\cdot\mathcal{L}_{\text{dice}}(\textbf{y}_l, \textbf{g}_l) \right) + \lambda_{cbDice} \cdot L_{cbDice}
\end{equation}

\section{Experiments and Results}

\subsection{Dataset and Evaluation Protocol}

\begin{table*}[!htbp]
\caption{Training protocols for the first and second stage models}
\label{table:training2nd}
\centering
\renewcommand{\arraystretch}{1.5}
\scriptsize
\begin{tabular}{ p{6cm} r }
\toprule[0.8pt]
\textbf{Training Parameter} & \textbf{Value} \\
\midrule[0.8pt]
Batch size & 2 \\
Patch size & 112$\times$112$\times$176 \\ 
Total epochs & 500 \\
Optimizer & SGD with nesterov momentum ($\mu=0.99$) \\ 
Initial learning rate (lr) & 0.01 \\ 
Training time & 15 hours/fold \\  
Number of model parameters & 210.6 M \\ 
\bottomrule[0.8pt]
\end{tabular}
\end{table*}

We used the dataset and evaluation framework provided by the AortaSeg24 Challenge \cite{imran2025multi}, organized as part of MICCAI 2024. The challenge focuses on multi-class segmentation of the aorta in computed tomography angiography (CTA), including 23 clinically relevant aortic branches and zones. The dataset consists of 100 annotated 3D CTA scans from patients with uncomplicated type B aortic dissection. Each volume was manually annotated by trained researchers and reviewed by an experienced vascular surgeon to ensure clinical accuracy. The annotations include major aortic branches (e.g., renal, iliac, and celiac arteries) and SVS/STS zones, following standard clinical guidelines. All volumes were resampled to an isotropic resolution of 1×1×1 mm³ for consistency.

Of the 100 scans, 50 were provided for training. The remaining 50 were split between validation and hidden test sets by the organizers. The use of external datasets was not permitted during model development.

\subsection{Implementation Details}
We implemented our model using the nnU-Net V2 framework \cite{isensee2021nnu,isensee2024nnu}. Data preprocessing and augmentation followed the default nnU-Net settings. The model was trained and evaluated on an NVIDIA RTX 3090 24GB GPU, using Python 3.10.9, PyTorch 2.2.2, and CUDA 12.1.

To evaluate segmentation performance, we used two standard metrics:
\begin{enumerate}
    \item \textbf{Dice Similarity Coefficient (DSC)} – quantifies volumetric overlap between predicted and reference segmentations.
    \item \textbf{Normalized Surface Distance (NSD)} – measures the boundary accuracy within a 2 mm tolerance.
\end{enumerate}

\subsection{Results on validation set}

\begin{figure}[h]
\centering
\includegraphics[width=\textwidth]{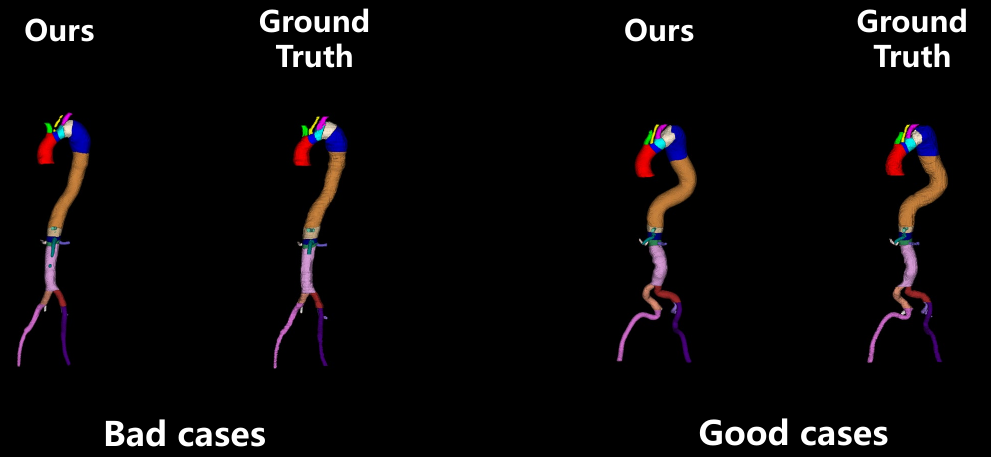}
\caption{Qualitative comparison of segmentation results. Left: Bad case showing discontinuities in small vessel categories (red arrows), likely due to limited training data for smaller structures. Right: Good case demonstrating accurate segmentation.}
\label{fig:aorta24_results}
\end{figure}

The hierarchical approach demonstrated faster training convergence and improved segmentation accuracy. As shown in Table~\ref{tab:epochs_datasets_results}, our model achieved a DSC of 70.15 and NSD of 65.87 on the validation set, outperforming the baseline nnU-Net ResEnc M model by 11.65 and 13.56 points, respectively. The clDice metric showed a slight decrease, indicating that while multi-class segmentation improved overall accuracy, it may have introduced some minor inconsistencies in small vessel segmentation.

\begin{table*}[!htbp]
\caption{Segmentation Performance on AortaSeg24 Dataset (Epoch 50)}
\label{tab:epochs_datasets_results}
\centering
\renewcommand{\arraystretch}{1.5}
\scriptsize
\begin{tabular}{ p{4.0cm} c c c }
\toprule[0.8pt]
\textbf{Model} & \textbf{Dice} & \textbf{NSD} & \textbf{clDice} \\
\midrule[0.8pt]
nnU-Net ResEnc M \cite{isensee2024nnu} & 58.50 & 52.31 & 94.29 \\
nnU-Net ResEnc M (w/ HC) & 70.15 & 65.87 & 93.26 \\
\midrule[0.8pt]
\textbf{Diff (w/ HC vs. no HC)} & \textcolor{blue}{+11.65} & \textcolor{blue}{+13.56} & \textcolor{red}{-1.03} \\
\bottomrule[0.8pt]
\end{tabular}
\end{table*}

\begin{table*}[!htbp]
\caption{Inference time comparison between one-stage and two-stage strategies.}
\label{tab1}
\centering
\renewcommand{\arraystretch}{1.5}
\scriptsize
\begin{tabular}{ p{4.5cm} c c }
\toprule[0.8pt]
\textbf{Inference Strategy} & \textbf{One Stage (default)} & \textbf{Two Stage ($m = 1, 2, 3, 4$)} \\
\midrule[0.8pt]
Time (s) & 115 & [22, 61, 98, 127] \\
\bottomrule[0.8pt]
\end{tabular}
\end{table*}

\begin{table}[h]
\caption{Quantitative evaluation on the AortaSeg24 testing dataset. The table reports average Dice and NSD scores across all 40 test cases for each anatomical region. Authors may expand this table to include ablation studies or additional baselines.}
\label{tab:results-testing}
\centering
\renewcommand{\arraystretch}{1.3} 
\scriptsize
\scalebox{0.8}{
\begin{tabular}{p{3.8cm} c c | c c}
\toprule[0.8pt]
\multirow{2}{*}{\textbf{Anatomical Region}}& \multicolumn{2}{c}{\textbf{Our Method}} & \multicolumn{2}{c}{\textbf{Baseline Method~\cite{imran2024cis}}} \\
& \textbf{Avg. DSC} & \textbf{Avg. NSD} & \textbf{Avg. DSC} & \textbf{Avg. NSD}\\
\midrule[0.8pt]
Zone 0 & 0.898 & 0.81 & 0.880 & 0.773\\
Innominate & 0.78 & 0.832 & 0.691 & 0.739\\
Zone 1 & 0.644 & 0.591 & 0.604 & 0.560\\
Left Common Carotid & 0.81 & 0.922 & 0.743 & 0.837\\
Zone 2 & 0.696 & 0.582 & 0.659 & 0.543\\
Left Subclavian Artery & 0.828 & 0.906 & 0.789 & 0.859\\
Zone 3 & 0.691 & 0.562 & 0.660 & 0.517\\
Zone 4 & 0.79 & 0.677 & 0.746 & 0.620\\
Zone 5 & 0.911 & 0.88 & 0.879 & 0.826\\
Zone 6 & 0.715 & 0.652 & 0.731 & 0.678\\
Celiac Artery & 0.675 & 0.844 & 0.568 & 0.728\\
Zone 7 & 0.703 & 0.672 & 0.699 & 0.660\\
SMA & 0.748 & 0.864 & 0.678 & 0.782\\
Zone 8 & 0.701 & 0.691 & 0.664 & 0.656\\
Right Renal Artery & 0.74 & 0.897 & 0.697 & 0.851\\
Left Renal Artery & 0.69 & 0.841 & 0.593 & 0.742\\
Zone 9 & 0.915 & 0.91 & 0.879 & 0.860\\
Right Common Iliac Artery & 0.866 & 0.921 & 0.800 & 0.840\\
Left Common Iliac Artery & 0.863 & 0.929 & 0.786 & 0.842\\
Right Internal Iliac Artery & 0.787 & 0.919 & 0.661 & 0.773\\
Left Internal Iliac Artery & 0.73 & 0.859 & 0.640 & 0.767\\
Right External Iliac Artery & 0.876 & 0.964 & 0.789 & 0.846\\
Left External Iliac Artery & 0.866 & 0.953 & 0.783 & 0.851\\
\midrule[0.8pt]
\textbf{Average} & \textbf{0.779} & \textbf{0.812} & \textbf{0.723} & \textbf{0.746}\\
\bottomrule[0.8pt]
\end{tabular}}
\end{table}

As shown in Table~\ref{tab1}, the two-stage inference strategy improves speed by up to five times compared to using only Model 2, depending on the expansion factor $m$.

The qualitative results are shown in Figure~\ref{fig:aorta24_results}. The bad case (left) exhibits fractures in some small vessel categories, which may be attributed to insufficient training data volume and suboptimal generalization capability for smaller anatomical structures. In contrast, the good case (right) demonstrates satisfactory segmentation performance.

\subsection{Results on Final Test Set}
In this section, discuss the performance of our model on the final test set, as compared to the baseline model, CIS-UNet~\cite{imran2024cis}. Table~\ref{tab:results-testing} summarizes the average Dice Similarity Coefficient (DSC) and Normalized Surface Distance (NSD) scores across 40 test images for each anatomical region. 

\section{Limitations and Future Work}
While our hierarchical semantic learning framework significantly improves segmentation accuracy and efficiency, several limitations remain. The model demonstrates suboptimal performance on small vessel categories, primarily due to the limited volume of training data available for these fine structures. Future work will focus on two key directions: (1) enhancing the model's generalization capabilities for smaller anatomical structures, and (2) extending the hierarchical semantic loss framework to a broader range of tasks.

\section{Conclusion}
We proposed a hierarchical semantic learning framework for multi-class aorta segmentation, which progressively learns anatomical semantics from simpler to more complex structures. The cbDice loss function ensures consistent segmentation across varying vessel diameters. The two-stage inference approach significantly reduces computation time, making the model suitable for real-time clinical applications.

\subsubsection{Acknowledgments} We thank all data contributors for making the medical images publicly available, and GrandChallenge for providing the challenge platform. This study was conducted as part of the MICCAI 2024 AortaSeg24 Challenge.

\subsubsection{Disclosure of Interests.} The authors have no competing interests to declare that are relevant to the content of this article.

\subsubsection{Ethical Compliance Statement.} All data used was publicly available and anonymized.
%
%
%
\bibliographystyle{splncs04}
\bibliography{Aorta_reference}

\end{document}